\documentclass[10pt,twocolumn,letterpaper]{article}

\usepackage{iccv}
\usepackage{times}
\usepackage{epsfig}
\usepackage{graphicx}
\usepackage{amsmath}
\usepackage{amssymb}
\usepackage[utf8]{inputenc} 
\usepackage[T1]{fontenc}    
\usepackage{CJKutf8}        
\usepackage{booktabs}
\usepackage{url}            
\usepackage{booktabs}       
\usepackage{amsfonts}       
\usepackage{nicefrac}       
\usepackage{microtype}      
\usepackage[dvipsnames]{xcolor}

\usepackage{multirow}
\usepackage{tabularx}
\usepackage{adjustbox}
\usepackage{wrapfig}
\usepackage{caption}        

\usepackage{makecell}
\usepackage{pifont}
\usepackage[abs]{overpic}

\usepackage{subfig}
\usepackage{graphicx}
\usepackage{tikz}
\usepackage{pgf, pgfsys, pgffor}
\usepackage{pgfplots}

\usepackage[accsupp]{axessibility} 

\usepackage[pagebackref=true,breaklinks=true,colorlinks,bookmarks=false]{hyperref}

\iccvfinalcopy 

\def\iccvPaperID{1666} 

\ificcvfinal\fi

\usepackage[capitalize]{cleveref}
\crefname{section}{Sec.}{Secs.}
\Crefname{section}{Section}{Sections}
\Crefname{table}{Table}{Tables}
\crefname{table}{Tab.}{Tabs.}
\Crefname{figure}{Figure}{Figures}
\crefname{figure}{Fig.}{Figs.}
\Crefname{equation}{Equation}{Equations}
\crefname{equation}{Eq.}{Eqs.}

\newcommand{\algname}{ProPainter}
\newcommand{\eat}[1]{}

\newcommand{\red}[1]{\textcolor{red}{#1}}
\newcommand{\blue}[1]{{\textcolor{blue}{#1}}}

\usepackage{hyperref}
\hypersetup{
colorlinks,
linkcolor={red},
citecolor={green},
urlcolor={purple}
}

\begin{document}
\begin{CJK}{UTF8}{gbsn}

\title{ProPainter: Improving Propagation and Transformer for Video Inpainting}

\author{Shangchen Zhou \quad Chongyi Li \quad Kelvin C.K. Chan \quad Chen Change Loy\\
S-Lab, Nanyang Technological University\\
{\tt\small \{s200094, chongyi.li, chan0899, ccloy\}@ntu.edu.sg}\\
{\tt\small \url{https://shangchenzhou.com/projects/ProPainter}}
}


\twocolumn[{%
\renewcommand\twocolumn[1][]{#1}%
\maketitle
\vspace{-8mm}
\begin{center}
    \centering
    \includegraphics[width=.99\linewidth]{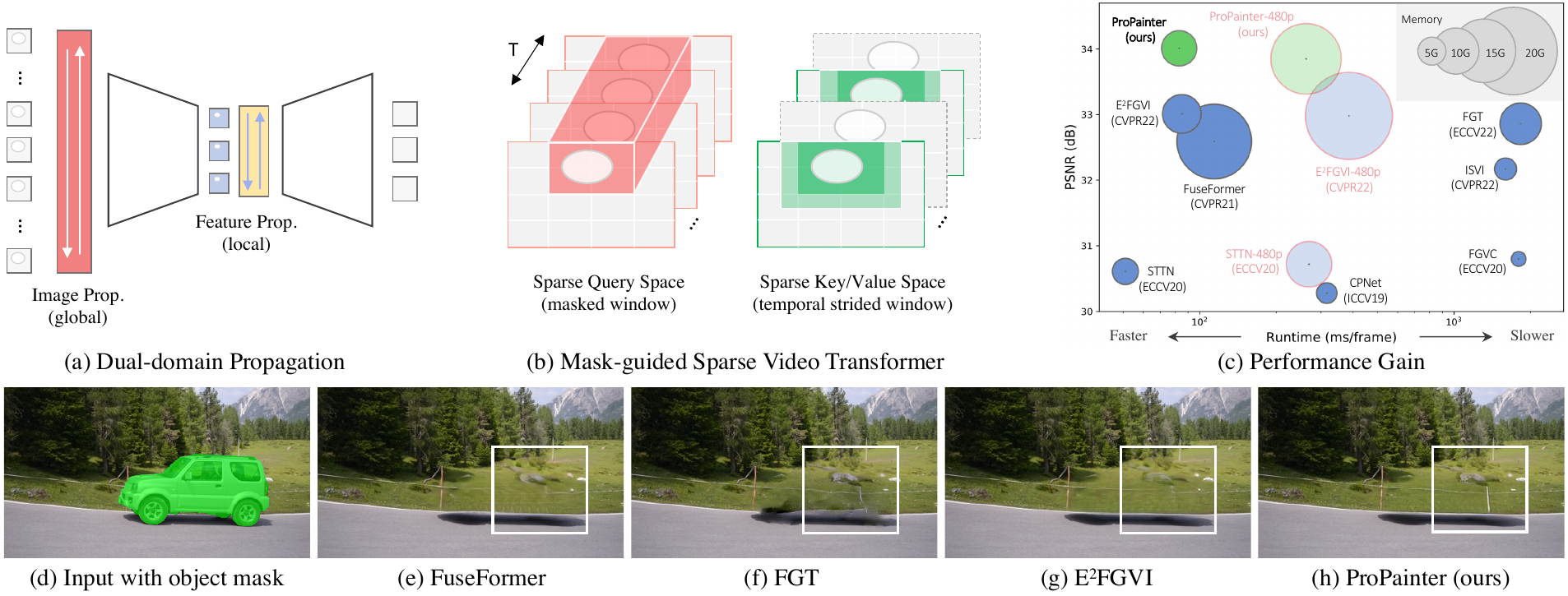}
    \vspace{-1mm}
    \captionof{figure}{
    (a) Dual-domain propagation enables more effective propagation due to its global and reliable nature. 
    (b) Mask-guided sparse video Transformer achieves high efficiency by discarding unnecessary and redundant windows.
    (c) ProPainter outperforms prior methods while maintaining efficiency.
    (d-h) In visual comparisons with FuseFormer~\cite{liu2021fuseformer}, FGT~\cite{zhang2022flow}, and E$^2$FGVI~\cite{li2022towards}, our ProPainter exhibits superiority in filling complete and rich textures.
    } \vspace{1mm}
    \label{fig:teaser}
\end{center}%
}]

\ificcvfinal\thispagestyle{empty}\fi

\begin{abstract}

Flow-based propagation and spatiotemporal Transformer are two mainstream mechanisms in video inpainting (VI). 
Despite the effectiveness of these components, they still suffer from some limitations that affect their performance. 
Previous propagation-based approaches are performed separately either in the image or feature domain. 
Global image propagation isolated from learning may cause spatial misalignment due to inaccurate optical flow. 
Moreover, memory or computational constraints limit the temporal range of feature propagation and video Transformer, preventing exploration of correspondence information from distant frames.
To address these issues, we propose an improved framework, called \textbf{ProPainter}, which involves enhanced \textbf{ProPa}gat\textbf{i}o\textbf{n} and an efficient \textbf{T}ransform\textbf{er}. 
Specifically, we introduce dual-domain propagation that combines the advantages of image and feature warping, exploiting global correspondences reliably. 
We also propose a mask-guided sparse video Transformer, which achieves high efficiency by discarding unnecessary and redundant tokens. 
With these components, ProPainter outperforms prior arts by a large margin of 1.46~dB in PSNR while maintaining appealing efficiency.

\end{abstract}


\vspace{-2.5mm}
\section{Introduction}
\label{sec:intro}
Video inpainting (VI) aims to fill gaps or missing regions in a video with visually consistent content while ensuring spatial and temporal coherence. 
This technique has broad applications, including video completion~\cite{gao2020flow}, object removal~\cite{ebdelli2015video, xu2019deep}, video restoration~\cite{tang2011video}, watermark, and logo removal~\cite{li2022towards}. VI is challenging because it requires establishing accurate correspondence across distant frames for information aggregation. 
To address this challenge, various mechanisms have been explored, such as 3D CNN~\cite{chang2019free, hu2020proposal}, video internal learning~\cite{zhang2019internal, ouyang2021internal}, flow-guided propagation~\cite{xu2019deep, gao2020flow, zhang2022inertia, zhang2022flow, li2022towards}, and video Transformer~\cite{liu2021fuseformer, zhang2022flow, li2022towards}. 
Among these mechanisms, flow-guided propagation and video Transformer have become mainstream choices for VI due to their promising performance.

Propagation-based methods in VI can be divided into two categories: image propagation and feature propagation. The former employs bidirectional global propagation in the image domain with a pre-completed flow field. While this approach can fill the majority of holes in a corrupted video, it requires an additional image or video inpainting network after propagation to hallucinate the remaining missing regions. This isolated two-step process can result in unpleasant artifacts and texture misalignment due to inaccurate flow, as shown in Figure~\ref{fig:teaser}(f). To address this issue, a recent approach called E$^2$FGVI~\cite{li2022towards} implements propagation in the feature domain, incorporating flow completion and content hallucination modules in an end-to-end framework. With the learnable warping module, the feature propagation module relieves the pressure of having inaccurate flow. However, E$^2$FGVI employs a downsampled flow field to match the spatial size of the feature domain, limiting the precision of spatial warping and the efficacy of propagation, potentially resulting in blurry results. Moreover, feature propagation can only be performed within a short range of video sequences due to memory and computational constraints, hindering propagation from distant frames and leading to missing texture, as shown in Figure~\ref{fig:teaser}(g).

Both image- and feature-based propagation have their pros and cons. In this study, we carefully revisit the VI problem and investigate the possibility of combining the strengths of both techniques. We demonstrate that with systematic redesigns and adaptation of best practices in the literature, we can achieve \textbf{dual-domain propagation}, as illustrated in Figure~\ref{fig:teaser}(a). To achieve reliable and efficient information propagation across a video, we identify several essential components:
\textit{i) Efficient GPU-based propagation with reliability check} -- Unlike previous methods that rely on complex and time-consuming CPU-centric operations, such as indexing flow trajectories, we perform global image propagation on GPU with flow consistency check. This implementation can be inserted at the beginning of the inpainting network and jointly trained with the other modules. Thus, subsequent modules are able to correct any propagation errors and benefit from the long-range correspondence information provided by the global propagation, resulting in a significant performance improvement.
\textit{ii) Improved feature propagation} -- Our implementation of feature propagation leverages flow-based deformable alignment~\cite{chan2022basicvsr++}, which improves robustness to occlusion and inaccurate flow completion compared to E$^2$FGVI~\cite{li2022towards}.
\textit{iii) Efficient flow completion} -- We design a highly efficient recurrent network to complete flows for dual-domain propagation, which is over 40 times ($\sim$192 fps\footnote{Tested on a single NVIDIA Tesla V100 GPU (32G).\label{v100}}) faster than SOTA method~\cite{zhang2022inertia} while maintaining comparable performance.
We demonstrate that these designs are essential to achieve efficient propagation of global and local information without texture misalignment or blurring in the filling results. An example is shown in Figure~\ref{fig:teaser}(h).

In addition to dual-domain propagation, we introduce an efficient \textbf{mask-guided sparse video Transformer} tailored for the VI task. The classic spatiotemporal Transformer is computationally intensive due to the quadratic number of interactions between video tokens, making it intractable for high-resolution and long temporal-length videos. For instance, contemporary Transformer-based methods, FuseFormer~\cite{liu2021fuseformer} and FGT~\cite{zhang2022flow}, are unable to handle 480p videos with a 32G GPU\textsuperscript{\ref{v100}} due to excessive memory demands. However, we observe that the inpainting mask usually covers only a small local region, such as the object area\footnote{Object regions account for only 13.6\% of the DAVIS~\cite{perazzi2016benchmark} dataset.}. Moreover, adjacent frames contain highly redundant textures. These observations suggest that spatiotemporal attention is unnecessary for most unmasked areas, and it is adequate to consider only alternating interval frames in attention computation. Motivated by these observations, we redesign the Transformer by discarding unnecessary and redundant windows in the query and key/value space, respectively, significantly reducing computational complexity and memory without compromising inpainting performance. 

The main contribution of this work is to provide a systematic study into the core problem of VI and offer a practical solution that is both effective and efficient. Propagating information in two distinct image and feature domains and combining them in a unified framework with fast GPU implementation is new for VI task. The mask-guided sparse video Transformer also offers practical insights into designing efficient spatiotemporal attention for VI task. Compared to the state-of-the-art methods, our model achieves superior performance with a large margin of 1.46 dB in PSNR, while also significantly reducing memory consumption.
%
%

\vspace{0mm}
\section{Related Work}
\vspace{0mm}
Numerous deep networks with different modules and propagation strategies have achieved significant success in video inpainting. These approaches can be broadly categorized into four categories:
\noindent {\bf 3D convolution.}
Earlier video inpainting networks typically employed 3D CNNs~\cite{chang2019free, wang2019video, hu2020proposal} or temporal shift~\cite{chang2019learnable} to aggregate spatiotemporal information. These methods often suffer from limited receptive fields in both temporal and spatial dimensions and misalignment between adjacent frames. As a result, they are less effective in exploring distant content.
\noindent {\bf Internal learning.}
To fully exploit content of a video, some studies~\cite{zhang2019internal, ouyang2021internal, ren2022dlformer} adopt internal learning to encode and memorize the appearance and motion of the video through deep networks. However, these methods require individual training for each test video, limiting their practical use.
\noindent {\bf Flow-guided propagation.}
Optical flow~\cite{kim2019deep, li2020short, zou2021progressive} and homography~\cite{lee2019copy, cai2022devit} are commonly used in video inpainting networks to align neighboring reference frames to enhance temporal coherence and aggregation. However, incomplete optical flow may not provide valid propagation for completing missing regions. To address this issue, recent flow-based methods~\cite{xu2019deep, gao2020flow, ke2021occlusion, zhang2022inertia, zhang2022flow} focus on first completing the flow and then use it as a guidance for pixel-domain propagation. This approach simplifies RGB pixel inpainting by completing a less complex flow field. However, this offline propagation is independent of the subsequent learnable refinement module, making it difficult to correct content distortion caused by inaccurate propagation. Inspired by flow-guided recurrent networks~\cite{chan2021basicvsr, chan2022basicvsr++}, Li et al.~\cite{li2022towards} proposed an end-to-end framework that jointly learns flow completion and feature propagation in the downsampled feature domain. However, downsampled flow reduces its ability to provide spatially precise warping. To overcome this limitation, we propose more faithful propagation by performing both pixel and feature propagation with flow consistency checks.
\noindent {\bf Video Transformer.}
Attention~\cite{lee2019copy, oh2019onion, hu2020proposal, li2020short} and Transformer~\cite{zeng2020learning, liu2021decoupled, liu2021fuseformer, cai2022devit, li2022towards, zhang2022flow} blocks adopt spatiotemporal attention to explore recurrent textures in a video. This enables them to retrieve and aggregate tokens with similar texture or context for filling in missing regions. 
Liu~\etal~\cite{liu2021fuseformer} present a fine-grained fusion Transformer based on the soft split and composition operations, which further boosts video inpainting performance.
However, these methods are computationally and memory intensive. To address this issue, some Transformers~\cite{liu2021decoupled, cai2022devit, zhang2022flow} decouple the spatiotemporal attention by performing spatial and temporal attention alternately, while others~\cite{li2022towards, zhang2022flow} adopt window-based Transformers~\cite{liu2021swin, yang2021focal} to reduce the spatial range for efficient video attention. However, these approaches still involve redundant or unnecessary tokens. Inspired by token pruning for adaptive attention~\cite{rao2021dynamicvit, yin2022vit, meng2022adavit, liang2022not, kong2022spvit} in high-level tasks, our study proposes a more efficient and faster video Transformer with sparse spatiotemporal attention and a largely reduced token space while maintaining inpainting performance.

Recent studies~\cite{li2020short, li2022towards, zhang2022flow} have demonstrated the effectiveness of combining flow-guided propagation with Transformer in VI. However, the high memory requirement of the Transformer limits the propagation range during both training and inference, severely hindering the ability to propagate temporally distant content. In this paper, we also adopt this combination strategy but propose a reliable propagation scheme, along with an efficient Transformer model that fully exploits the benefits of long-range propagation and attention. This results in superior inpainting performance while maintaining computational efficiency.
%
%

\section{Methodology}
\label{sec:method}
%
\begin{figure*}[th]
\begin{center}
    \includegraphics[width=\linewidth]{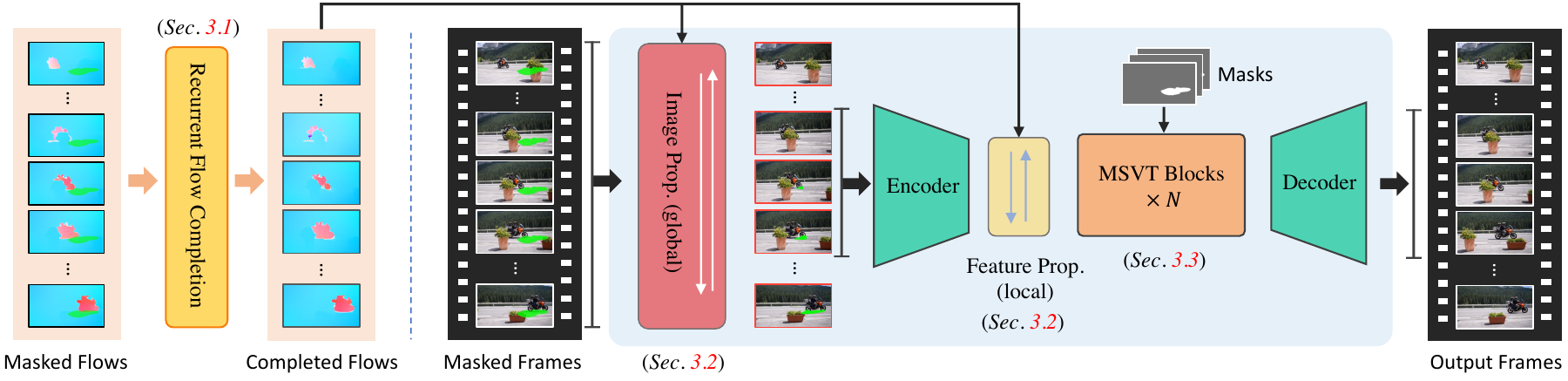}
    \vspace{-3mm}
    \caption{
    ProPainter comprises three key components: recurrent flow completion, dual-domain propagation, and mask-guided sparse Transformer. First, we employ a highly efficient recurrent flow completion network to complete the corrupted flow fields. We then perform propagation in both image and feature domains, which are jointly trained. This approach enables us to explore correspondences from both global and local temporal frames, resulting in more reliable and effective propagation. The subsequent mask-guided sparse Transformer blocks refine the propagated features using spatiotemporal attention, aided by a sparse strategy that considers only a subset of the tokens. This enhances efficiency and reduces memory consumption, while maintaining performance.    
    }
    \vspace{-6mm}
\label{fig:overview}
\end{center}
\end{figure*}
Given a masked video sequence $X=\{X_{t} \in \mathbb{R}^{ H\times W\times 3}\}_{t=1}^T$, which has a sequence length of $T$, along with corresponding mask sequence $M=\{M_{t} \in \mathbb{R}^{ H\times W\times 1}\}_{t=1}^T$, the objective of video inpainting is to generate visually consistent and coherent content within the corrupted or missing regions. 
ProPainter, as shown in Figure~\ref{fig:overview}, is composed of three key components: {Recurrent Flow Completion} (RFC), {Dual-Domain Propagation} (DDP), and {Mask-guided Sparse Video Transformer} (MSVT).
Before feeding the sequence into ProPainter, we extract the forward and backward optical flows, denoted as $F^f=\{F^f_t = F_{t\rightarrow t+1} \in \mathbb{R}^{ H\times W\times 2}\}_{t=1}^{T-1}$ and $F^b=\{F^b_t = F_{t+1\rightarrow t} \in \mathbb{R}^{ H\times W\times 2}\}_{t=1}^{T-1}$ from a given video $X$.
We first use RFC to complete the corrupted flow fields. 
Guided by the completed flows, we then perform global image propagation and local feature propagation sequentially. 
Finally, we employ multiple MSVT blocks to refine propagation features and a decoder to reconstruct the final video sequence $\hat{Y}=\{\hat{Y}_{t} \in \mathbb{R}^{ H\times W\times 3}\}_{t=1}^T$. 
We introduce the specific design of each component below.
\subsection{Recurrent Flow Completion}
\label{sec:rfc}
Pre-trained flow completion modules are commonly used in video inpainting networks~\cite{xu2019deep, gao2020flow, zhang2022inertia, zhang2022flow}. The rationale behind this approach is that it is simpler to complete missing flow than to directly fill in complex RGB content~\cite{xu2019deep}. Furthermore, using completed flow to propagate pixels reduces the pressure of video inpainting and better maintains temporal coherence. E$^2$FGVI~\cite{li2022towards} proposes to insert the flow completion module into an end-to-end framework, which simplifies the inpainting pipeline. However, flow completion modules that are learned together with inpainting-oriented losses can result in a suboptimal learning process and less accurate completed flow. Moreover, the downsampled flow may limit the precision of spatial warping and the efficacy of propagation, which can result in blurred and incomplete filling content, as shown in Figure~\ref{fig:teaser}(g). Therefore, an independent flow completion model is not only important but also necessary for video inpainting.

To maintain temporal coherence while completing flows, previous methods~\cite{xu2019deep, zhang2022flow} adopt sliding-window-based networks to aggregate optical flow information from adjacent frames, which are highly correlated. However, these methods can be computationally expensive as repeated inferences are required in the overlapping frames. 
To improve efficiency and enhance flow coherence further, we adopt a recurrent network~\cite{chan2021basicvsr, chan2022basicvsr++} for flow completion, which provides precise optical flow fields for subsequent propagation modules.

We complete forward and backward flows using the same process, thus we denote $F^f$ and $F^b$ as $F$ for simplicity. 
We first encode the flows $F_t$ into a downsampled feature $f_t$ with a downsampling ratio of 8. 
Next, we employ deformable alignment~\cite{chan2022basicvsr++} that is based on deformable convolution (DCN)~\cite{dai2017deformable, zhu2019deformable}, to bidirectionally propagate the flow information from nearby frames for flow completion. 
For simplicity, we only describe the backward propagation process here. 
Taking the concatenated feature $c(f_t, \hat{f}_{t+1})$, where $\hat{f}_{t+1}$ is the propagation feature of the \texttt{t+1}-th frame, as input a lightweight network with a stack of convolutions is employed to compute DCN offsets $o_{t\rightarrow t+1}$ and modulation masks $m_{t\rightarrow t+1}$. DCN alignment propagation can be expressed as: 
\begin{equation}
\hat{f_t} = \mathcal{R} \big(\mathcal{D} (\hat{f}_{t+1}; o_{t\rightarrow t+1}, m_{t\rightarrow t+1}), f_t\big),
\label{eq:dcn_wo_flow}
\end{equation}
where $\mathcal{D}(\cdot)$ denotes deformable convolution, and $\mathcal{R}(\cdot)$ denotes the convolution layers that fuse the aligned and current features. In this way, information of ($t+1$)-th flow can be adaptively transferred to the current $t$-th flow. 
Finally, a decoder is used to reconstruct the completed flows $\hat{F_t}$. 
For clarity, an illustration of deformable alignment is provided in the supplementary material.
\subsection{Dual-domain Propagation}
After completing the flow, we perform global and local propagation in the image and feature domains, respectively. We employ distinct alignment operations and strategies for each domain. Both domains involve bidirectional propagation in the forward and backward directions. Here, we elaborate on the backward propagation since the forward propagation follows the same process.

\noindent {\bf Image propagation.}
To maintain efficiency and simplicity, we adopt flow-based warping for image propagation, along with a simple reliability check strategy.
This process does not involve any learnable operation. 
In the case of a video sequence $X$ with binary masks $M$ (a pixel with value 1 represents masked region) and completed flows $\hat{F}$, we first check the validity of completed flow based on forward-backward consistency error~\cite{xu2019deep, gao2020flow}: 
\begin{equation}
\mathcal{E}_{t\rightarrow t+1}\big(p\big) = \Big\| \hat{F}_{t\rightarrow t+1}\big(p\big) +\hat{F}_{t+1\rightarrow t}\big(p+\hat{F}_{t\rightarrow t+1}(p)\big)\Big\|_2^2,
\label{eq:flow_check}
\end{equation}
where $p$ denotes a pixel position of the current frame. Only pixels with a small consistency error will be propagated, i.e., $C_1: \mathcal{E}_{t\rightarrow t+1}(p) < \epsilon$, where $\epsilon$ is a threshold and set to 5. 
Furthermore, we only consider the masked areas of the current frame $X_t$ that needs to be filled, i.e., $C_2: M_t(p)=1$, and we only propagate the unmasked areas from neighboring frame $X_{t+1}$, i.e., $C_3: M_{t+1}(p+\hat{F}_{t\rightarrow t+1}(p))=0$.  By enforcing the three constraints, a reliable propagation area $A_r$ is identified as:
\begin{equation}
{A_r}\big(p\big) =
\begin{cases}
1  & \text{if } p \in C_1 \cap C_2 \cap C_3 , \\
0  & \text{otherwise}. \\
\end{cases}
\label{eq:reliable_region}
\end{equation}
The process of image propagation is expressed as:
\begin{equation}
{\hat{X}}_{t} = \mathcal{W}\big(X_{t+1}, \hat{F}_{t\rightarrow t+1}\big) * A_r +  X_t * \big(1-A_r\big),
\label{eq:image_warp}
\end{equation}
where $\mathcal{W}(\cdot)$ denotes warping operation. 
To ensure continuous propagation, we promptly update the mask $M_t$ of the current frame and convert the propagated area to the unmasked status by updating masks via $\hat{M}_t = M_t - A_r$. 
After global image propagation, we obtain a partially filled video sequence $\hat{X}$, which greatly eases the learning process for subsequent modules.

\begin{figure}[t]
\begin{center}
    \includegraphics[width=.99\linewidth]{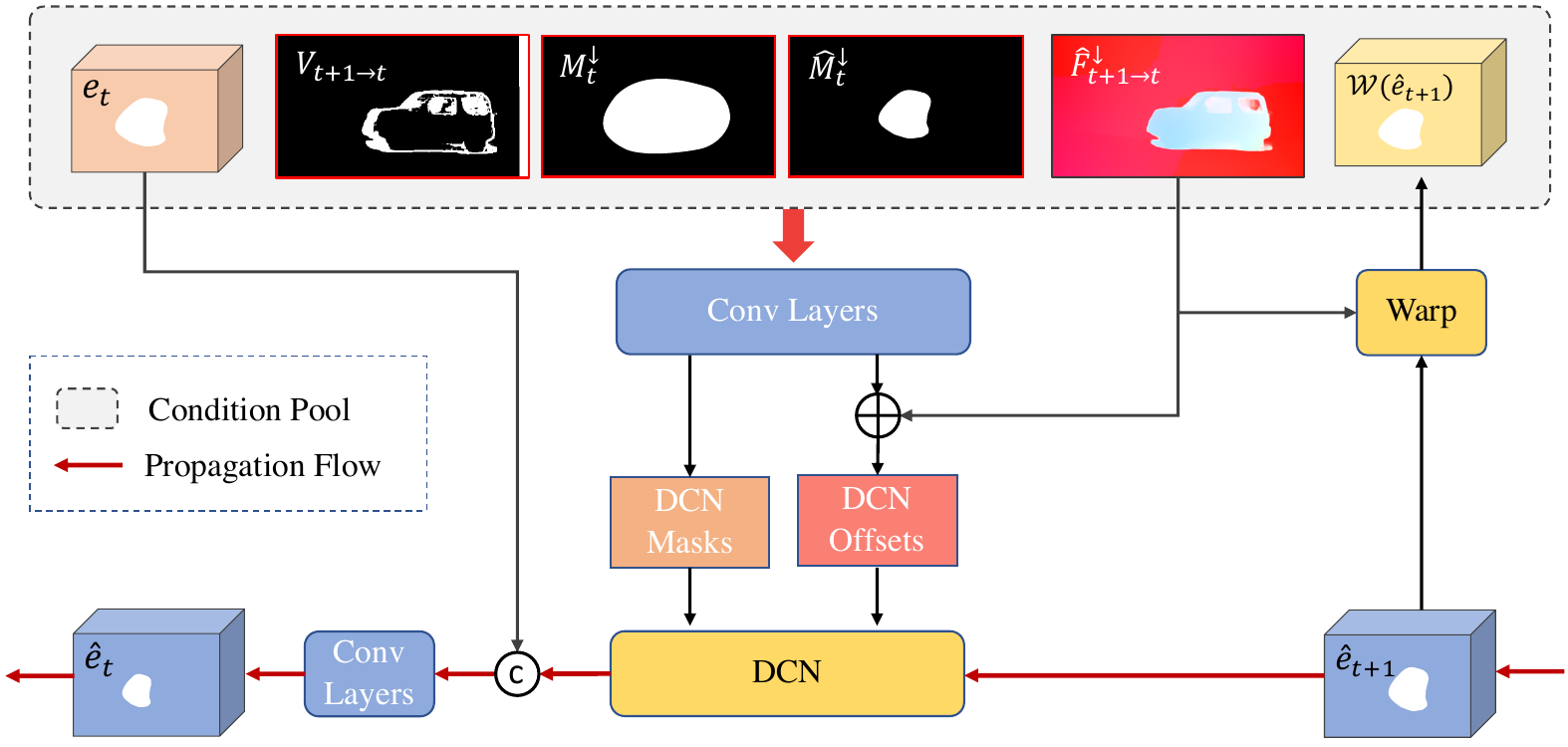}
    \caption{Flow-guided deformable alignment is effective by taking reliable completed flows and mask-aware conditions. We concatenate the validated flow map, original mask, and updated mask into conditions to produce DCN offsets (residue to optical flow). A DCN is then applied to align the propagation feature from the previous frame. Finally, a CNN block is employed to fuse the current and aligned features, achieving the propagation feature of the current frame.}
\label{fig:feat_dcn}
\end{center}
\vspace{-3mm}
\end{figure}
\noindent {\bf Feature propagation.}
We use an image encoder with the same structure as previous works~\cite{liu2021fuseformer, li2022towards} to extract features from a local sequence $\hat{X}_{t=1}^{T_l}$, denoted as $\{e_{t} \in \mathbb{R}^{ \frac{H}{4}\times \frac{W}{4}\times C}\}_{t=1}^{T_l}$.
Similar to E$^2$FGVI~\cite{li2022towards}, we also adopt flow-guided deformable alignment module~\cite{chan2022basicvsr++} for feature propagation, which has demonstrated remarkable benefits in various low-level video tasks~\cite{chan2022generalization, chan2022investigating, zhou2019spatio}.
Unlike the deformable alignment used in Sec.~\ref{sec:rfc} that directly learns DCN offsets, flow-guided deformable alignment employs the completed flow as a base offset and refines it by learning offset residue.
However, our design differs from E$^2$FGVI in that we offer richer conditions for learning DCN offsets. 
As illustrated in Figure~\ref{fig:feat_dcn}, apart from the current feature $e_{t}$, warped propagation feature $\mathcal{W}(\hat{e}_{t+1}, \hat{F}_{t\rightarrow t+1}^{\downarrow})$, and completed flows $\hat{F}_{t\rightarrow t+1}^{\downarrow}$, we additionally introduce the flow valid map $V_{t+1\rightarrow t}$ calculated by consistency check (Eq.~\ref{eq:flow_check}), as well as the original mask $M_t^{\downarrow}$, and updated mask $\hat{M}_t^{\downarrow}$ after image propagation. 
With these conditions, a stack of convolutions is employed to predict the DCN offset residue $\widetilde{o}_{t\rightarrow t+1}$ and modulation masks $m_{t\rightarrow t+1}$. 
The flow-guided DCN alignment propagation is expressed as:
\begin{equation}
\hat{e_t} = \mathcal{R} \big(\mathcal{D} (\hat{e}_{t+1}; \hat{F}_{t\rightarrow t+1}^{\downarrow} + \widetilde{o}_{t\rightarrow t+1}, m_{t\rightarrow t+1}), f_t\big),
\label{eq:dcn_w_flow}
\end{equation}
where $\downarrow$ denotes downsampling. 
The improved reliability of flow and the additional awareness of mask as a condition make our flow-guided deformable alignment module more stable to learn than previous designs~\cite{chan2022basicvsr++, li2022towards}. The current step is able to focus more on truly challenging regions where flow is invalid and former image propagation is unreliable.

\subsection{Mask-Guided Sparse Video Transformer}
\begin{figure}[th]
\begin{center}
    \includegraphics[width=.99\linewidth]{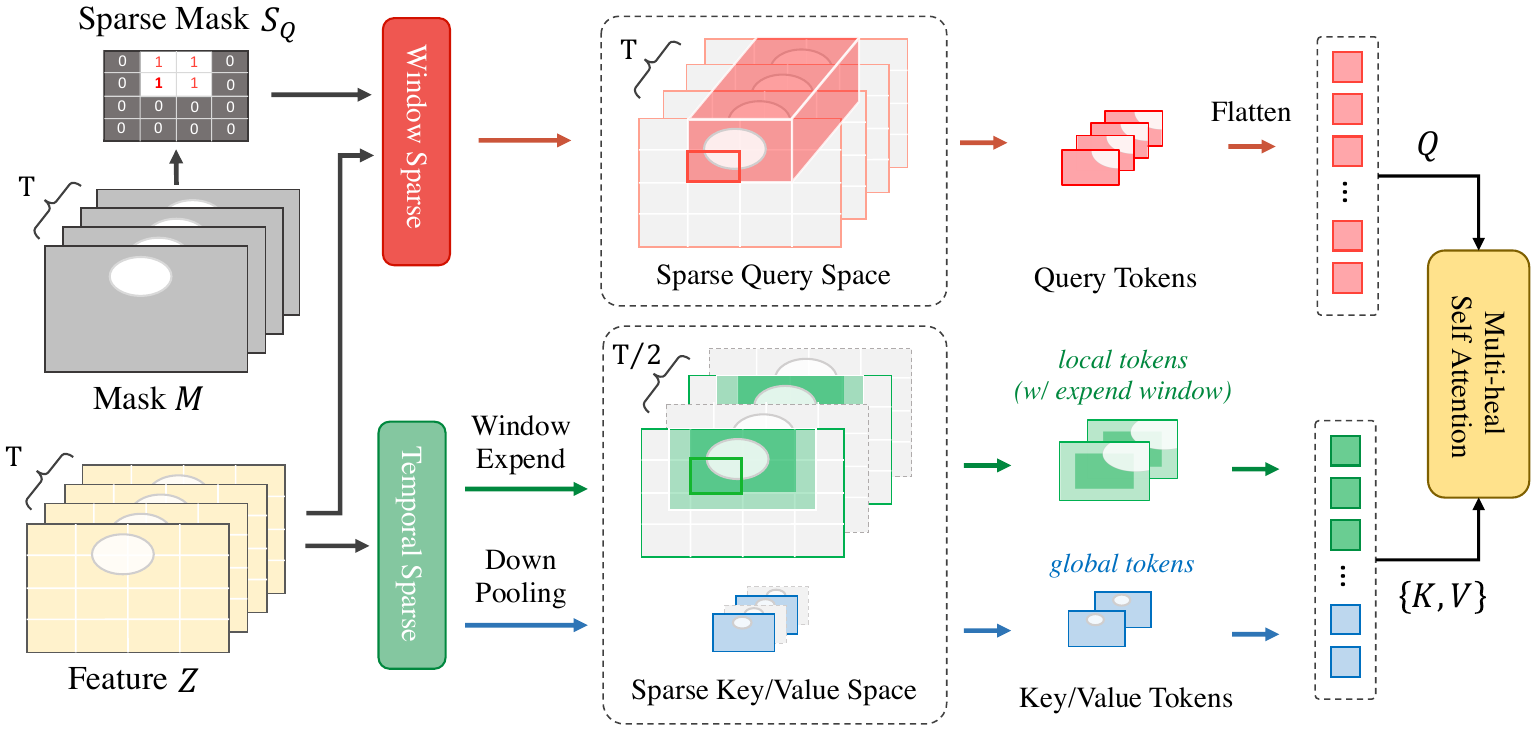}
    \caption{Mask-guided sparse video Transformer. To reduce computational complexity and memory usage, our mask-guided sparse Transformer filters out unnecessary and redundant windows in the query and key/value space, respectively, before applying self-attention. To enlarge spatial interrelation range, we also adopt the window expand strategy~\cite{yang2021focal} and pooling global tokens~\cite{zhang2022flow, li2022towards}.}
\label{fig:feat_prop}
\end{center}
\vspace{-3mm}
\end{figure}
While video Transformers have achieved excellent performance in video inpainting, they can be computationally and memory intensive, posing a challenge to their practical application. E$^2$FGVI and FGT have addressed this issue by using window-based Transformer blocks, but they still have some efficiency limitations. To overcome this, we propose a novel sparse video Transformer that builds on the window-based approach. 
Given a video sequence feature $E_{l}\in \mathbb{R}^{T_l \times \frac{H}{4}\times \frac{W}{4}\times C}$, we use the soft split operation~\cite{liu2021fuseformer} to generate patch embeddings $Z\in \mathbb{R}^{T_l \times M\times N\times C_z}$. We partition $Z$ into $m\times n$ non-overlapping windows, resulting in partitioned features $Z_w\in \mathbb{R}^{T_l \times m\times n \times h\times w\times C_z}$, where $m\times n$ and $h\times w$ are the number and size of the windows, respectively. 
We obtain the query $Q$, key $K$, and value $V$ from $Z_w$ through linear layers. We design sparse strategies for both query and key/value spaces separately. 
Note that we also apply the window expand strategy~\cite{liu2021fuseformer} and integrate global tokens~\cite{zhang2022flow} into key and value, enabling us to use a small window size of $5\times 9$ in our experiments. 
We omit them from the following discussion since they do not affect our sparse strategy designs.
\begin{table*}[t]
    \begin{center}
    \caption{
        Quantitative comparisons on YouTube-VOS~\cite{xu2018youtube} and DAVIS~\cite{perazzi2016benchmark} datasets. The best and second performances are marked in \red{\underline{red}} and \blue{{blue}}, respectively.
        $E^*_{warp}$ denotes $E_{warp}~(\times 10^{-3})$.
        All methods are evaluated following their default settings.
        Since DFVI, FGVC, ISVI, and FGT involve several CPU processes, their FLOPs cannot be accurately projected.
        }
    \label{tab:comparison}
    \vspace{-2mm}
    \renewcommand{\arraystretch}{1.15}
    \renewcommand{\tabcolsep}{2.05mm}
    \scalebox{0.9}{
    \begin{tabular}{l|c|c|c|c||c|c|c|c||c|c}
    \hline
    \cline{1-11}
    & \multicolumn{8}{c||}{Accuracy} & \multicolumn{2}{c}{Efficiency} \\
    \cline{2-11}
    & \multicolumn{4}{c||}{YouTube-VOS} & \multicolumn{4}{c||}{DAVIS} & FLOPs  & Runtime  \\
    \cline{1-9}
    Models & PSNR $\uparrow$ & SSIM $\uparrow$ & VFID $\downarrow$ &  $E^*_{warp} \downarrow$ & PSNR $\uparrow$ & SSIM $\uparrow$ & VFID $\downarrow$ & $E^*_{warp} \downarrow$  & (10 frames) & (s/frame) \\
    \cline{1-11}
    \hline
    DFVI~\cite{xu2019deep}              & 29.16 & 0.9429 & 0.066 & 1.651 & 28.81 & 0.9404 & 0.187 & 1.596 & -     & 0.837 \\
    \hline
    CPNet~\cite{lee2019copy}               & 31.58 & 0.9607 & 0.071 & 1.622 & 30.28 & 0.9521 & 0.182 & 1.521 & 1407G & 0.316 \\
    \hline
    FGVC~\cite{gao2020flow}             & 29.67 & 0.9403 & 0.064 & 1.163 & 30.80 & 0.9497 & 0.165 & 1.571 & -     & 1.795 \\
    \hline
    STTN~\cite{zeng2020learning}              & 32.34 & 0.9655 & 0.053 & 1.061 & 30.61 & 0.9560 & 0.149 & 1.438 & 1315G & 0.051 \\
    \hline
    TSAM~\cite{zou2021progressive}   & 30.22 & 0.9468 & 0.070 & 1.014    & 30.67 & 0.9548 & 0.146  & \blue{{1.235}}    & 1001G  & 0.068 \\
    \hline
    FuseFormer~\cite{liu2021fuseformer}   & 33.32 & 0.9681 & 0.053 & 1.053 & 32.59 & 0.9701 & 0.137 & 1.349 & 1025G  & 0.114 \\
    \hline
    ISVI~\cite{zhang2022inertia} & 30.34 & 0.9458 & 0.077 & \blue{1.008}    & 32.17 & 0.9588 & 0.189  & 1.291     & -  & 1.594 \\
    \hline
    FGT~\cite{zhang2022flow} & 32.17 & 0.9599 & 0.054 & 1.025    & 32.86 & 0.9650 & 0.129  & 1.323    & -  & 1.828 \\
    \hline
    E$^2$FGVI~\cite{li2022towards}  & \blue{{33.71}} & \blue{{0.9700}} & \blue{{0.046}} & {{1.013}} & \blue{{33.01}} & \blue{{0.9721}} & \blue{{0.116}}  & 1.289 & 986G & 0.085 \\
    \hline
    \hline
    \algname~(Ours) & \red{\underline{34.43}} & \red{\underline{0.9735}} & \red{\underline{0.042}} & \red{\underline{0.974}} &      \red{\underline{34.47}} & \red{\underline{0.9776}} & \red{\underline{0.098}}  & \red{\underline{1.187}}     & 808G  &  0.083\\
    \hline
    \cline{1-11}
    \end{tabular}
    }
    \end{center}
    \vspace{-4mm}
\end{table*}
\noindent {\bf Sparse Query Space.} 
We observe that mask regions often occupy only a small area of the video, such as in the case of object removal in the DAVIS~\cite{perazzi2016benchmark} dataset, where the proportion of object regions is only 13.6\%. This indicates that spatiotemporal attention may not be necessary for all query windows. 
To exploit this observation, we selectively apply attention to query windows that intersect with the mask regions. 
Specifically, we first use nearest neighbor interpolation to downsample the mask sequence $M\in \mathbb{R}^{T_l \times H\times W}$ to $M^{\downarrow}\in \mathbb{R}^{T_l \times m\times n}$, where $m\times n$ is the number of non-overlapping windows after partitioning. We then sum it up in the temporal dimension and obtain sparse mask $S_{Q} \in \mathbb{R}^{m\times n}$ for query cubes following the equation:
\begin{equation}
S_{Q} = Clip~\Big(\sum \nolimits_{t=1}^{T_l} M^{\downarrow}_t, ~1\Big),
\label{eq:dcn_w_flow}
\end{equation}
where $Clip$ represents a clipping function that set $S_{Q}$ to 1 if $\sum \nolimits_{t=1}^{T_l} M^{\downarrow}_t > 0$. In other words, if the query cube at a window $(i,j)$ has never contained any mask region in the past frames, then $S_{Q}(i,j) = 0$, indicating that spatiotemporal attention within this window can be skipped.  
\noindent {\bf Sparse Key/Value Space.} 
Due to the highly redundant and repetitive textures in adjacent frames, it is unnecessary to include all frames as key/value tokens in each Transformer block. Instead, we will only include strided temporal frames alternately, with a temporal stride of 2 in our design. That is, in each odd-numbered Transformer block, only odd-number frames are activated to participate in self-attention with their key and value, while even-number blocks include only even-number frames. By doing so, the key and value space is reduced by half, effectively reducing the computation and memory cost of the Transformer module.
After filtering out unnecessary and redundant windows based on our sparse strategy, we perform self-attention on the remaining windows to extract refined features. These features are then gathered using a soft composition operation~\cite{liu2021fuseformer} for subsequent modules.
Experimental results suggest that our design significantly reduces the computational cost of video Transformers while maintaining performance for video inpainting.

\subsection{Training Objectives}

\noindent {\bf Flow Completion.} We utilize L1 loss as the reconstruction loss and a second-order smoothness constraint on the flow field~\cite{meister2018unflow} to promote the collinearity of neighboring flows and thus enhance the smoothness of the completed flow field.
\noindent {\bf Video Inpainting.}
We adopt L1 loss as the reconstruction loss for all pixels. 
To enhance the realistic and temporal consistency of video inpainting results, we also employ an adversarial loss that is measured using a T-PatchGAN~\cite{chang2019free} discriminator.
The details and formulation of these losses are provided in the supplementary material.
%
%

\vspace{0mm}
\section{Experiments}
\label{sec:exp}
\noindent {\bf Datasets.}
We use the training set of YouTube-VOS~\cite{xu2018youtube} with 3471 video sequences to train our networks. 
Two widely-used  test sets are adopted for evaluation:  YouTube-VOS~\cite{xu2018youtube} and DAVIS~\cite{perazzi2016benchmark}, which consist of 508 and 90 video sequences, respectively. 
For the DAVIS test set, following FuseFormer~\cite{liu2021fuseformer} and E$^2$FGVI~\cite{li2022towards}, we use 50 video clips for evaluations. 
During training, we follow~\cite{kim2019deep, lee2019copy, liu2021fuseformer, li2022towards} and generate stationary and object masks in a random fashion to simulate the masks in video completion and object removal tasks. 
As for evaluation, we adopt the stationary masks provided in~\cite{li2022towards} to calculate quantitative scores, and the object masks are extracted from their segmentation labels for qualitative comparisons. Video frames are sized to $432\times 240$ for training and evaluation.
\begin{figure*}[t]
\begin{center}
    \includegraphics[width=.99\linewidth]{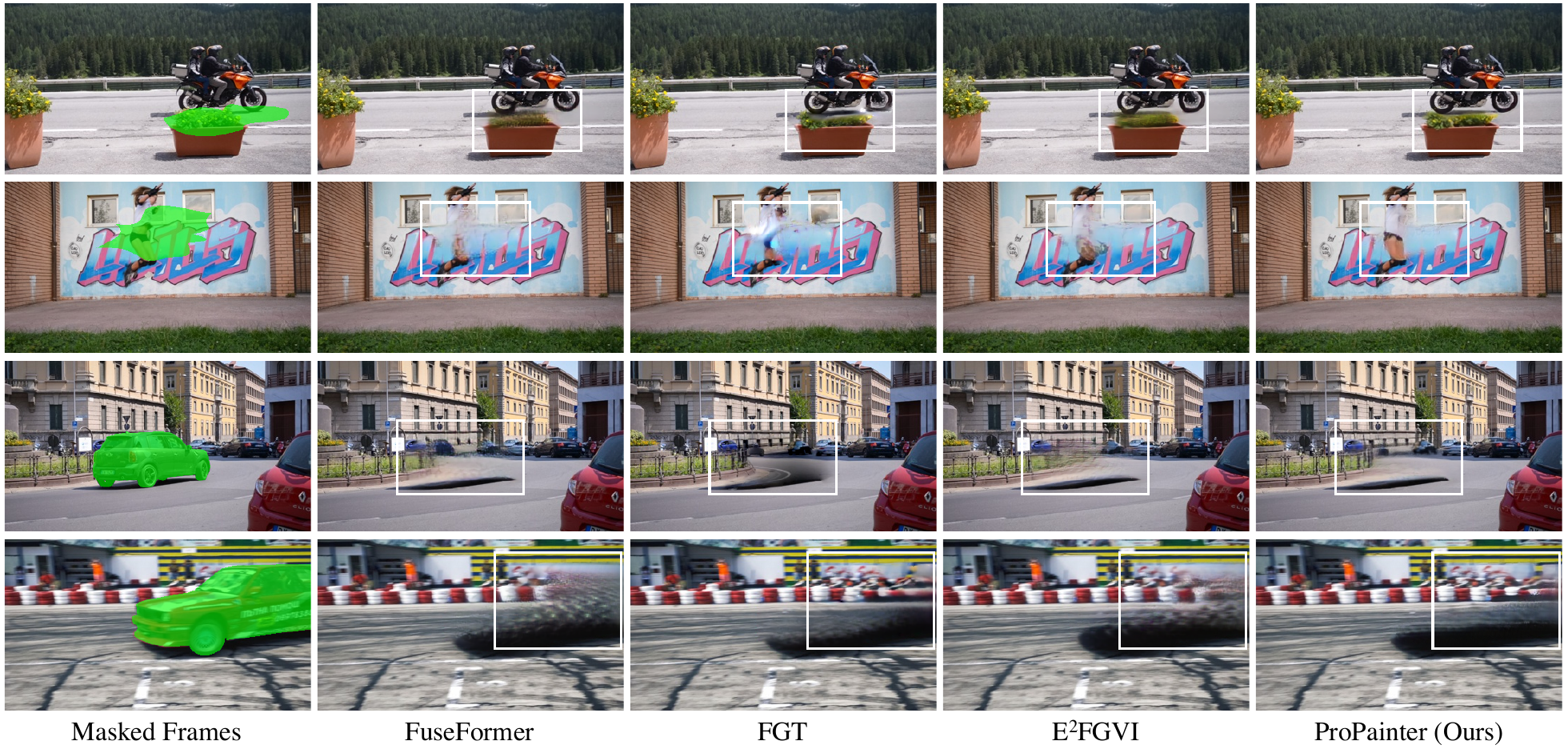}
    \vspace{-1mm}
    \caption{Qualitative comparisons on both video completion and object removal. Our ProPainter exhibits superiority in producing complete and faithful textures, resulting in enhanced spatiotemporal coherence for video inpainting.
    }
\label{fig:visual_compare}
\end{center}
\vspace{-5mm}
\end{figure*}
\noindent {\bf Training Details and Metrics.}
We use RAFT~\cite{teed2020raft} to extract optical flow in our approach. 
For training the RFC network, we set the flow sequence length to 10 and perform deformable propagation on feature maps that are downsampled by a factor of 8 for faster processing. 
We adopt 8 Transformer blocks for the inpainting modules and use a local video sequence of length 10. 
The Transformer window size is $5\times 9$, and the extended size is half of the window size. 
We train both the RFC and inpainting modules using the Adam~\cite{kingma2014adam} optimizer with a batch size of 8, setting the initial learning rate to $10^{-4}$ and running 700k iterations\footnote{We set 450k training iterations for ablation study.} for each. We implement our method using the PyTorch framework and train it on 8 NVIDIA Tesla V100 (32G) GPUs.
We employ the widely used PSNR and SSIM metrics~\cite{wang2004image} to evaluate the reconstruction performance and VFID~\cite{wang2018video} scores to measure the perceptual similarity between input videos and outputs, as used in recent video inpainting studies~\cite{liu2021fuseformer, li2022towards}. 
Additionally, we report the flow warping error $E_{warp}$~\cite{lai2018learning} to assess the temporal consistency and smoothness of the resulting video sequences.
\subsection{Comparisons}
\noindent {\bf Quantitative Evaluation.}
We compare ProPainter with nine state-of-the-art methods  including DFVI~\cite{xu2019deep},  CPNet~\cite{lee2019copy}, FGVC~\cite{gao2020flow}, STTN~\cite{zeng2020learning}, TSAM~\cite{zou2021progressive}, Fuseformer~\cite{liu2021fuseformer}, ISVI~\cite{zhang2022inertia}, FGT~\cite{zhang2022flow}, and E$^2$FGVI~\cite{li2022towards} on both YouTube-VOS~\cite{xu2018youtube} and DAVIS~\cite{perazzi2016benchmark}. Thanks to the efficient design, ProPainter uses a temporal length of 20 for inference.
Table~\ref{tab:comparison} shows that ProPainter outperforms other methods in all quantitative metrics, especially on the DAVIS dataset, where our method surpasses the state-of-the-art method by 1.14 dB in PSNR. The results suggest that our method has superior inpainting capability, enabling it to produce higher-quality, faithful, and seamless videos.
\noindent {\bf Qualitative Evaluation.}
For the visual comparison, we compare our method with FuseFormer~\cite{liu2021fuseformer}, FGT~\cite{zhang2022flow}, and E$^2$FGVI~\cite{li2022towards}, which are representative methods of Transformer-, image propagation-, and feature propagation-based approaches, respectively. Figure~\ref{fig:visual_compare} presents four comparison results for video completion and object removal.
Our method uses dual-domain propagation to ensure reliable and long-range propagation. It completes missing regions with coherence and clear contents, while other compared methods tend to fail or produce unpleasant inpainting results such as texture distortions and black hazy region in FGT~\cite{zhang2022flow} results, as well as artifacts in FuseFormer~\cite{liu2021fuseformer} and E$^2$FGVI~\cite{li2022towards}. 
\noindent {\bf Efficiency Comparison.}
Table~\ref{tab:comparison} presents the efficiency comparisons between all methods in terms of FLOPs and running time. The FLOPs of all methods are computed based on a temporal length of 10. We consider all learnable modules (including the recurrent flow completion) in our ProPainter to calculate the FLOPs. The running time records the time of all processes in each method, including inpainting, as well as flow calculation and flow completion if involved. To keep efficiency, we use only five iterations of the RAFT network to calculate optical flow.
\begin{table}[t]
\caption{Comparisons of flow completion networks. Our network offers a dual benefit with high accuracy and efficiency.}
\centering
\vspace{-2mm}
\resizebox{0.99\linewidth}{!} {
\renewcommand{\arraystretch}{1.15}
\renewcommand{\tabcolsep}{2.05mm}
\begin{tabular}{lccccc}
\toprule
EPE $\downarrow$ & DFVI~\cite{xu2019deep} & FGVC~\cite{gao2020flow}  & FGT~\cite{zhang2022flow}  & ISVI~\cite{zhang2022inertia}   &  Ours  \\ 
\midrule
YouTube-VOS & 0.046  & 0.032  & 0.021  & \bf{0.019}   & 0.020   \\ \midrule
DAVIS & 0.107   & 0.082  & 0.052    & \bf{0.051}    & \bf{0.051}      \\ \midrule\midrule
Runtime (s/frame) & 0.130   & 1.125  & 0.312    & 0.231    & \bf{0.005} \\
\bottomrule
\end{tabular}
}
\label{tab:flownet_compare}
\vspace{-5mm}
\end{table}
\noindent {\bf Flow Completion Comparisons.}
We compare our recurrent flow network with previous approaches~\cite{xu2019deep, gao2020flow, zhang2022inertia} on both YouTube-VOS and DAVIS datasets. Table~\ref{tab:flownet_compare} presents the end-point-error (EPE) of flow completion and running time of each method. Our recurrent network offers a dual benefit with high accuracy and efficiency. Compared to previous methods, our network is approximately 40 times faster while maintaining a comparable flow completion accuracy to the state-of-the-art methods.

\begin{table*}[t]
\caption{Ablation study of dual-main propagation and sparse Transformer.}
\centering
\vspace{-2mm}
\resizebox{0.99\linewidth}{!} {
\begin{tabular}{lcccc|c|c}
\toprule
Exp. & (a) w/o Img Prop. &  (b) w/ Img Prop. in FGVC & (c) w/o Feat Prop.  & (d) w/ Feat Prop. in E$^2$FGVI & (f) Full Tokens & ProPainter\\ 
\midrule
PSNR &  33.05 &  32.91  &   33.17  & 33.94         & {34.18} & 34.15 \\ \midrule
SSIM & 0.9724   &  0.9687 &  0.9732 & 0.9756     & {0.9765} & 0.9764  \\ \bottomrule
\end{tabular}
}
\label{tab:ablation}
\vspace{0mm}
\end{table*}
\begin{figure*}[t]
\begin{center}
    \vspace{-1mm}
    \includegraphics[width=.99\linewidth]{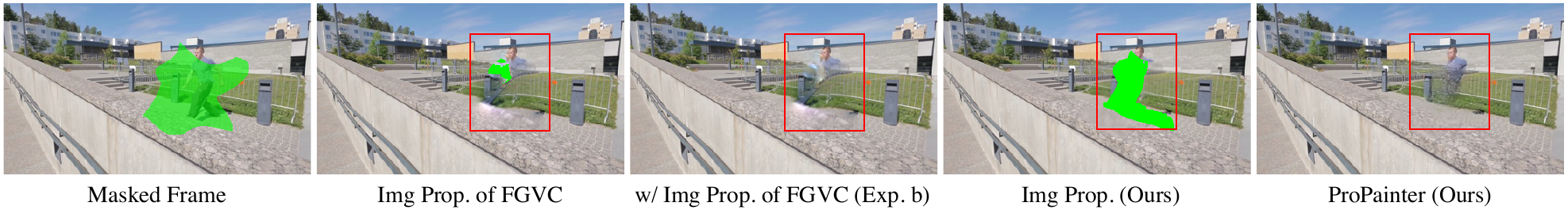}
    \vspace{-2mm}
    \caption{Visual comparison on image propagation methods of FGVC~\cite{gao2020flow} and ours.}
\label{fig:img_prop_compare}
\end{center}
\vspace{-5.mm}
\end{figure*}
\subsection{Ablation Study}
\begin{figure}[t]
  \centering
  \vspace{-1.5mm}
   \includegraphics[width=\linewidth]{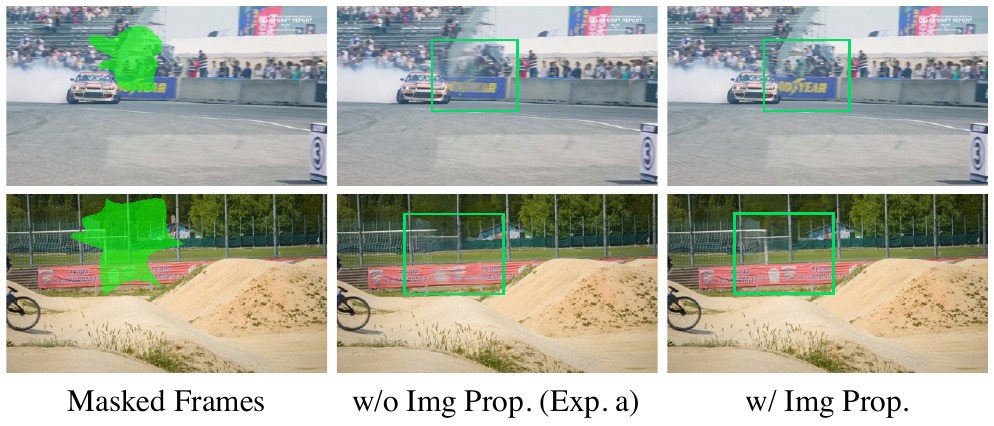}
   \vspace{-6mm}
   \caption{Comparison of w/ and w/o image propagation.}
   \vspace{-5mm}
   \label{fig:img_prop_ablation}
\end{figure}
\noindent {\bf Effectiveness of Image Propagation.} 
Table~\ref{tab:ablation} shows that Exp. (a) experiences a significant performance drop when image propagation is removed. Moreover, the model's propagation ability is reduced without image propagation, as presented in Figure~\ref{fig:img_prop_ablation}, causing it to fail to complete missing content with details. To verify the effectiveness of our reliability check strategy in image propagation, we replaced our design with the FGVC image propagation module in Exp. (b) (without retraining), resulting in a noticeable decrease in PSNR. This is because the FGVC image propagation method is prone to being affected by incorrect optical flow, leading to severe texture distortion that subsequent modules cannot correct. Our model can effectively aware and stop unreliable propagation areas using a simple reliability check via Eq.\ref{eq:flow_check}, and generate more faithful inpainting results.

\noindent {\bf Effectiveness of Feature Propagation.}
Similarly, we observe a slight decrease in performance by either removing feature propagation, \ie, Exp. (c), or replacing it with the Feature propagation of E$^2$FGVI, \ie, Exp. (d), indicating the effectiveness of the feature propagation modules and our reliability mask-aware conditions. This suggests that our design, which learns reliable DCN offsets in the feature domain, can further complement and enhance the propagation ability in the image domain.

\noindent {\bf Effectiveness of Sparse Transformer.}
In theory, our strategy of using masks to guide sparsity only eliminates redundant and unnecessary tokens (windows), while preserving essential information. This means that there should be no adverse effect on performance. To confirm this, we conducted Exp. (d), comparing our approach to standard self-attention without sparse filtering. Our results indicate that our sparse Transformer block performs almost as well as the standard one, indicating that it can achieve high efficiency without sacrificing performance.

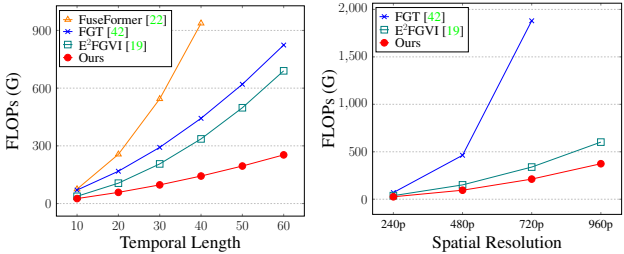
\begin{figure}[t]
\vspace{-5mm}
\hspace{-3mm}
  \centering
  \subfloat{
    \resizebox{.483\linewidth}{!} {
      \begin{tikzpicture}[every plot/.append style={ultra thick}]
        \begin{axis} [
          width = \textwidth,
          xlabel = {Temporal Length},
          xlabel style = {yshift=-0.65cm, font=\fontsize{32}{32}\selectfont},
          ylabel = {FLOPs (G)},
          ylabel style = {yshift=1.5cm, font=\fontsize{32}{32}\selectfont},
          log basis x = {2},
          xtick = {10, 20, 30, 40, 50, 60},
          xticklabel style = {yshift=-0.2cm, font=\fontsize{32}{32}\selectfont},
          ytick = {0, 300, 600, 900},
          yticklabel style = {xshift=-0.2cm, font=\fontsize{24}{24}\selectfont},
          legend style = {at={(0.465, 0.978)}, column sep=0.3cm, inner sep=0.2cm, font=\fontsize{24}{24}\selectfont},
          legend cell align = left,
          ymajorgrids = true,
          grid style = dashed,
        ]

        \addplot[
          color=orange,
          mark=triangle,
          mark size=6pt
        ]
        coordinates {
          (10, 75.1) (20, 256) (30, 544) (40, 937)
        };
        \addlegendentry{FuseFormer~\cite{liu2021fuseformer}};

        \addplot[
          color=blue,
          mark=x,
          mark size=6pt
        ]
        coordinates {
          (10, 70) (20, 168) (30, 292) (40, 443) (50, 620) (60, 824)
        };
        \addlegendentry{FGT~\cite{zhang2022flow}};

        \addplot[
          color=teal,
          mark=square,
          mark size=6pt
        ]
        coordinates {
          (10, 37.65) (20, 106) (30, 206) (40, 336) (50, 498) (60, 690)
        };
        \addlegendentry{E$^2$FGVI~\cite{li2022towards}};

        \addplot[
          color=red,
          mark=*,
          mark size=6pt
        ]
        coordinates {
          (10, 25.77) (20, 58.1) (30, 97) (40, 143) (50, 195) (60, 253)
        };
        \addlegendentry{Ours};
        \end{axis}
        \label{fig:gflops-length}
      \end{tikzpicture}
    }
   }
   \subfloat{
    \resizebox{.485\linewidth}{!} {
      \begin{tikzpicture}[every plot/.append style={ultra thick}]
        \begin{axis} [
          width = \textwidth,
          xlabel = {Spatial Resolution},
          xlabel style = {yshift=-0.75cm, font=\fontsize{32}{32}\selectfont},
          ylabel = {FLOPs (G)},
          ylabel style = {yshift=1.5cm, font=\fontsize{32}{32}\selectfont},
          log basis x = {2},
          xtick = {240, 480, 720, 960},
          xticklabels = {240p, 480p, 720p, 960p},
          xticklabel style = {yshift=-0.2cm, font=\fontsize{22}{22}\selectfont},
          ytick = {0, 500, 1000, 1500, 2000},
          yticklabel style = {xshift=-0.2cm, font=\fontsize{24}{24}\selectfont},
          legend style = {at={(0.385, 0.98)}, column sep=0.3cm, inner sep=0.2cm, font=\fontsize{24}{24}\selectfont},
          legend cell align = left,
          ymajorgrids = true,
          grid style = dashed,
        ]

        \addplot[
          color=blue,
          mark=x,
          mark size=6pt
        ]
        coordinates {
          (240, 70) (480, 463) (720, 1880) 
        };
        \addlegendentry{FGT~\cite{zhang2022flow}};

        \addplot[
          color=teal,
          mark=square,
          mark size=6pt
        ]
        coordinates {
          (240, 37.65) (480, 151) (720, 339)  (960, 602) 
        };
        \addlegendentry{E$^2$FGVI~\cite{li2022towards}};

        \addplot[
          color=red,
          mark=*,
          mark size=6pt
        ]
        coordinates {
          (240, 25.77) (480, 95) (720, 212)  (960, 374) 
        };
        \addlegendentry{Ours};
        \end{axis}
        \label{fig:gflops-resolution}
      \end{tikzpicture}
    }
   }
   \vspace{-1mm}
   \caption{FLOPs cures of different Transformer blocks.}
   \vspace{-3mm}
   \label{fig:flops_compare}
\end{figure}

\noindent {\bf Efficiency of Sparse Transformer.}
In Figure~\ref{fig:flops_compare}, we compare the FLOPs of different Transformer blocks with respect to temporal length and spatial resolution, including those used in FuseFormer~\cite{liu2021fuseformer}, FGT~\cite{zhang2022flow}, and E$^2$FGVI~\cite{li2022towards}. We use a mask with a missing region ratio of 1/6 (higher than the average object ratio of 13.6\% in DAVIS) to calculate the FLOPs of our mask-guided sparse Transformer. The curves indicate that the efficiency advantage of our sparse Transformer becomes more prominent as the temporal length and video resolution increase, indicating great potential for developing longer-range spatiotemporal attention and applying it to larger resolution videos.

%
%

\section{Conclusion}
\vspace{0mm}
This study introduces a novel and improved video inpainting framework called ProPainter. It incorporates an enhanced dual-domain propagation and an efficient mask-guided sparse video Transformer. Thanks to the two modules, our ProPainter exhibits reliable and precise propagation capabilities over long distances, significantly improving the performance of video inpainting while maintaining high efficiency in terms of running time and computational complexity. We believe that the designs in ProPainter will provide valuable insights to the video inpainting community.

\vspace{2mm}
\noindent{\bf Acknowledgement.} This study is supported under the RIE2020 Industry Alignment Fund Industry Collaboration Projects (IAF-ICP) Funding Initiative, as well as cash and in-kind contribution from the industry partner(s).

\clearpage

{\small
\bibliographystyle{ieee_fullname}
\bibliography{iccv23_bib}
}

\clearpage

\onecolumn 
\renewcommand\thesection{\Alph{section}}
\setcounter{section}{0}
\begin{center}
	\Large\textbf{{ProPainter: Improving Propagation and Transformer for Video Inpainting}}\\
	\vspace{15pt}
	\Large{-- Supplementary Materials --} \\
	\vspace{25pt}
\end{center}

In this supplementary materials, we provide additional details, further discussions, and more results to supplement the main paper. In Sec.~\ref{sec:arch}, we present the architecture details and loss functions of our proposed ProPainter.
In Sec.~\ref{sec:discussion}, we provide in-depth analysis of the performance improvement achieved by our method and highlight its advantages.
Sec.~\ref{sec:result} contains more quantitative evaluations and visual comparisons.

\section{Architecture and Loss Details}
\label{sec:arch}
\subsection{Architecture}
\begin{figure}[ht]
  \centering
   \includegraphics[width=0.42\linewidth]{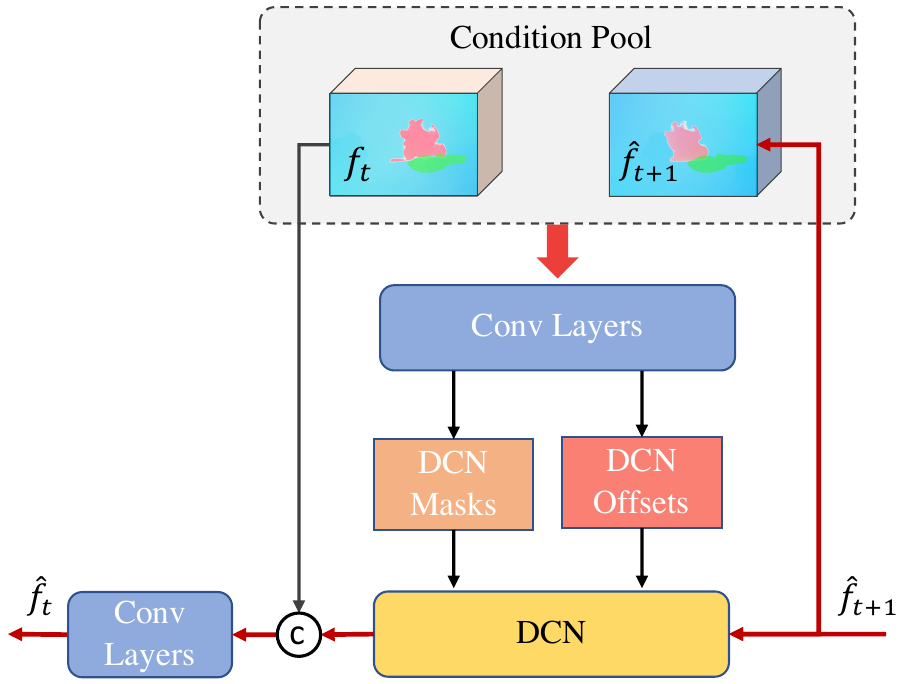}
   \caption{An illustration of deformable alignment module that is adopted in the recurrent flow completion network.}
   \label{fig:dcn_align_woflow}
\end{figure}

Our network adopts two distinct deformable alignment modules in the recurrent flow completion network (RFC) and feature propagation, respectively. To provide further clarity, we present a detailed illustration of the former alignment module (w/o flow guided) in Figure~\ref{fig:dcn_align_woflow}, which can be easily compared with the latter alignment module (w/ flow guided) depicted in Figure~{\color{red}3} of the main paper. There are two main differences between the two modules: 1) different condition pools were employed to predict the parameters of the deformable convolutional networks (DCN); 2) the former predicts the DCN offsets directly, while the latter uses optical flow as the base offset of DCN and predicts the residual offsets to the flow fields.

%

\subsection{Loss Functions}
\noindent\textbf{Loss Functions of RFC network.}
To train the recurrent flow completion network (RFC), we utilize two losses. The first one is the reconstruction loss that is applied to both valid and invalid regions, as depicted in the following equation:
\begin{align}
\vspace{-2mm}
\label{eq:rfc_loss1}
\mathcal{L}_{rec}^{flow} = \frac{\left\| M_t \odot (\hat{F}_t - F_t) \right\|_1}{\left\| M_t \right\|_1} + \frac{\left\| (1-M_t) \odot (\hat{F}_t - F_t) \right\|_1}{\left\| 1-M_t \right\|_1},
\end{align}
where $\odot$ denotes the dot product. The second one is second-order smooth loss~\cite{meister2018unflow} that encourages the smooth and coherent completed flow fields, which is a critical property for the subsequent propagation modules.
The loss can be expressed as:
\begin{align}
\vspace{-2mm}
\label{eq:rfc_loss2}
\mathcal{L}_{smooth}^{flow} = \left\| \bigtriangleup \hat{F}_t \right\|_1,
\end{align}
where $\bigtriangleup$ denotes the divergence operator. The overall loss function of RFC is: $\mathcal{L}^{flow} = \alpha_{1} \mathcal{L}_{rec}^{flow} + \alpha_{2} \mathcal{L}_{smooth}^{flow}$, where we set $\alpha_{1} = 1, \alpha_{2} = 0.5$ in our experiments.
\noindent\textbf{Loss Functions of ProPainter.}
Our ProPainter is trained using two types of loss. For reconstruction loss, we use L1 loss to measure the distance between output video sequence $\hat{Y}$ and  ground-truth one $Y$: 
\begin{align}
\label{eq:rec_loss}
\mathcal{L}_{rec} = \left\| \hat{Y}_t - Y_t \right\|_1.
\end{align}
Furthermore, we introduce an adversarial training procedure with a T-PatchGAN based discriminator $D$~\cite{chang2019free} to enhance the quality and coherence of generated videos by differentiating between real and reconstructed videos:
\begin{align}
\label{eq:d_loss}
\mathcal{L}_{D} = \mathbb{E}_{\text{Y}} \Big[\text{log} D(\text{Y})\Big] + \mathbb{E}_{\hat{\text{Y}}} \Big[1-\text{log} D(\hat{\text{Y}})\Big].
\end{align}
For the generator, the GAN loss is formulated as: 
\begin{align}
\label{eq:g_loss}
\mathcal{L}_{G} = - \mathbb{E}_{\hat{\text{Y}}} \Big[\text{log} D(\hat{\text{Y}})\Big].
\end{align}
Thus, the objective of ProPainter learning is: $\mathcal{L}^{inpaint} = \lambda_{1} \mathcal{L}_{rec} + \lambda_{2} \mathcal{L}_{G}$, where we set $\lambda_{1} = 1, \lambda_{2} = 0.01$.

\section{More Discussions}
\label{sec:discussion}
As indicated in Table~{\color{red}1} of the  main paper, our proposed ProPainter outperforms the state-of-the-art networks by a large margin on all quantitative metrics, especially on the DAVIS~\cite{perazzi2016benchmark} dataset. In this section, we explore  the primary factor that contributes to these remarkable performance gains and discuss the situations in which our approach has a competitive edge.
\subsection{Factor Behind Improved Performance}
\begin{figure}[ht]
  \centering
  \vspace{-2mm}
   \includegraphics[width=\linewidth]{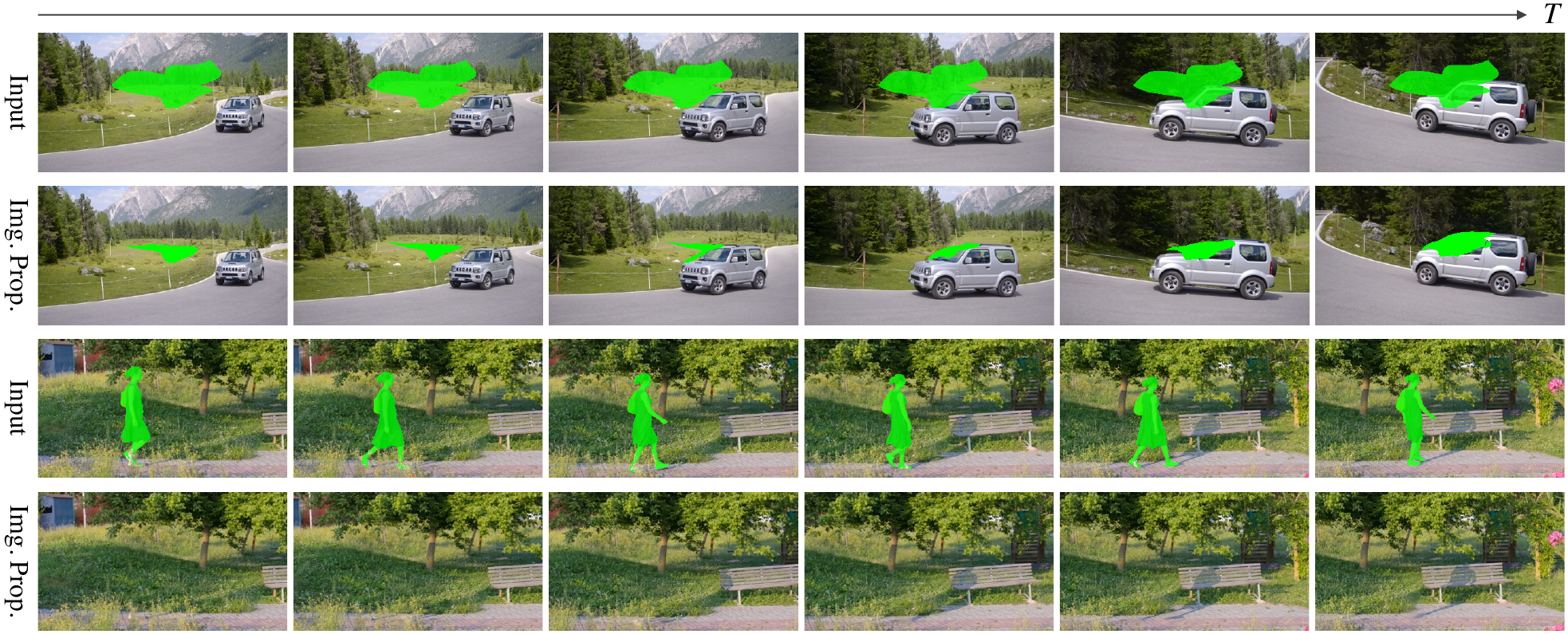}
   \vspace{-5mm}
   \caption{The initial results and updated masks after our global image propagation. Image propagation shows effective to fill most or entire masked regions, which significantly alleviates the learning difficulty experienced by video inpainting networks.}
   \vspace{-2mm}
   \label{fig:img_prop_results}
\end{figure}

Our proposed ProPainter benefits greatly from global image propagation, which significantly reduces the difficulty of learning for subsequent modules. As shown in Figure~\ref{fig:img_prop_results}, image propagation has filled the majority of the masks and even entirely completed masked regions. This means that modules following image propagation only need to refine and complement the completed contents of image propagation instead of learning the entire inpainting process. Our method differs from previous approaches~\cite{gao2020flow, zhang2022flow, zhang2022inertia} in several aspects: 1) In contrast to earlier image propagation methods that are independent of network training, which prevent the network from correcting propagation errors, our proposed image propagation is involved in the model training, enabling subsequent models to fix any texture misalignment or artifacts caused by image propagation; 2) We employ a more reliable propagation strategy, which is compared in Figure~{\color{red}6} of the main paper; 3) Unlike previous methods that are implemented on the CPU and involved some complex and time-consuming processes, such as indexing pixel-wise flow trajectories and Poisson blending, we implement a more efficient image propagation with GPU acceleration.

\subsection{Motion Distribution}
\begin{figure}[ht]
  \centering
   \includegraphics[width=0.43\linewidth]{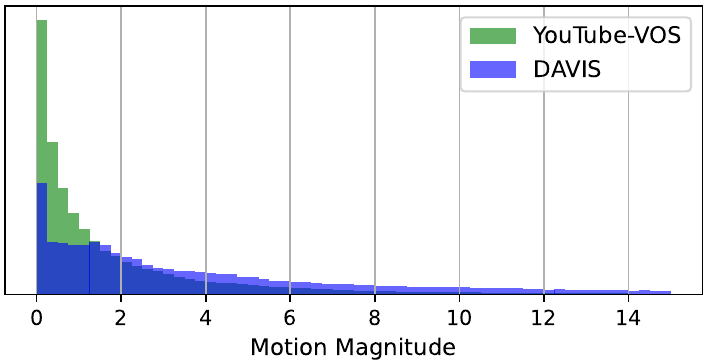}
   \caption{Motion magnitude distribution on YouTube-VOS~\cite{xu2018youtube} and DAVIS~\cite{perazzi2016benchmark} datasets.}
   \label{fig:motion_dist}
\end{figure}

In the main paper, Table~{\color{red}1} shows that ProPainter's performance improvement is more noticeable on the DAVIS~\cite{perazzi2016benchmark} dataset than on the YouTube-VOS~\cite{xu2018youtube} dataset. Our ablation study and analysis in the main paper attribute the performance gains primarily to the design of dual-domain propagation, which relies on motion flow fields to propagate information across videos. However, we have observed that many videos in the YouTube-VOS dataset have almost stationary scenes without motion, which limits the effectiveness of our dual-domain propagation module. Moreover, we have analyzed the motion magnitude distribution on both datasets and found that the YouTube-VOS dataset contains a greater proportion of regions with small motion, as presented in Figure \ref{fig:motion_dist}.

\section{More Results}
\label{sec:result}
\subsection{Quantitative Evaluation on 480p Videos}
Table~\ref{tab:compare_480p} presents a quantitative comparison on the DAVIS~\cite{perazzi2016benchmark} dataset with 480p ($864\times 480$) videos. The comparison only includes STTN~\cite{zeng2020learning} and E$^2$FGVI~\cite{li2022towards}, since other methods require memory demands exceeding 32G (such as TSAM~\cite{zou2021progressive}, FuseFormer~\cite{liu2021fuseformer}, and FGT~\cite{zhang2022flow}) or excessively long time for inference on a 480p video. Runtimes are measured on an NVIDIA Tesla V100 (32G) GPU. This comparison suggests that our method exhibits benefits in terms of both accuracy and efficiency even at a high resolution.
\begin{table}[h]
\caption{Quantitative comparisons on DAVIS~\cite{perazzi2016benchmark} dataset with 480p ($864\times 480$) videos.}
\centering
\vspace{-1mm}
\resizebox{0.54\linewidth}{!} {
\renewcommand{\arraystretch}{1.15}
\renewcommand{\tabcolsep}{2.05mm}
\begin{tabular}{lcccc}
\toprule
 & PSNR $\uparrow$ & SSIM $\uparrow$  & VFID $\downarrow$  & Runtime (s/frame) $\downarrow$ \\ 
\midrule
STTN~\cite{zeng2020learning} & 30.72  & 0.9534  & 0.055  & {0.262} \\ \midrule
E$^2$FGVI~\cite{li2022towards} & 32.98   & 0.9693  & 0.041   & {0.332}    \\ \midrule\midrule
ProPainter (Ours) & \bf{33.81}   & \bf{0.9739}  & \bf{0.035}    & \bf{0.249} \\
\bottomrule
\end{tabular}
}
\label{tab:compare_480p}
\end{table}
\clearpage
\subsection{Qualitative Comparisons on Flow Completion}
\begin{figure}[ht]
  \centering
   \includegraphics[width=0.99\linewidth]{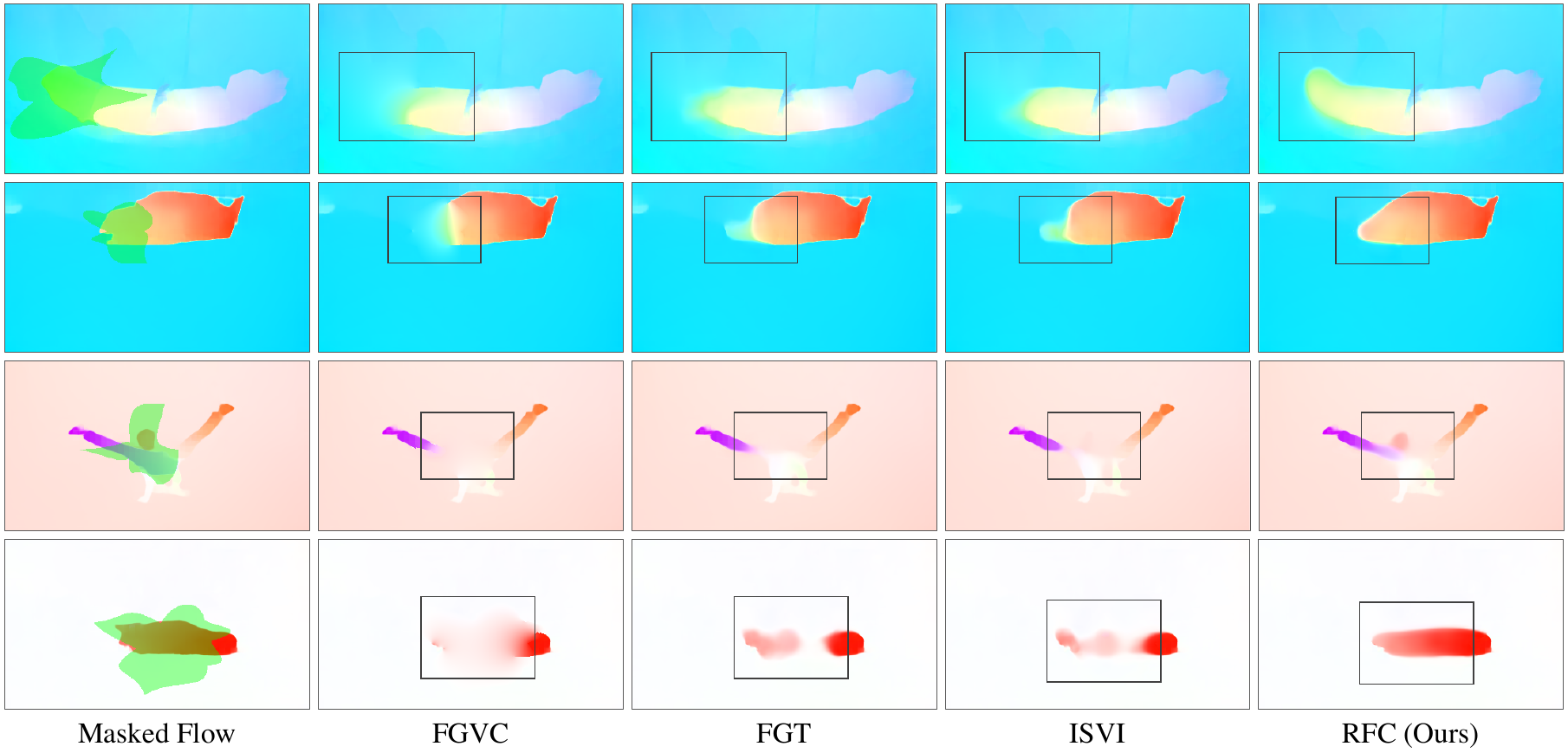}
   \caption{Qualitative comparisons of flow completion. Our recurrent flow completion network exhibits superiority in generating complete and faithful flow fields, thereby facilitating more precise and reliable propagation for ProPainter.}
   \label{fig:flow_compare}
\end{figure}

In Figure~\ref{fig:flow_compare}, we provide a visual comparison of flow completion performance between our recurrent flow completion network and previous methods, including FGVC~\cite{gao2020flow}, FGT~\cite{zhang2022flow}, and ISVI~\cite{zhang2022inertia}. The results show that our recurrent flow completion network outperforms other methods in producing complete and accurate flow fields. As a result, the subsequent dual-domain propagation module can rely more on accurate optical flows, leading to a more reliable and precise propagation in later stages.

\clearpage
\subsection{Qualitative Comparisons}
\begin{figure}[ht]
  \centering
  \vspace{-1mm}
   \includegraphics[width=0.99\linewidth]{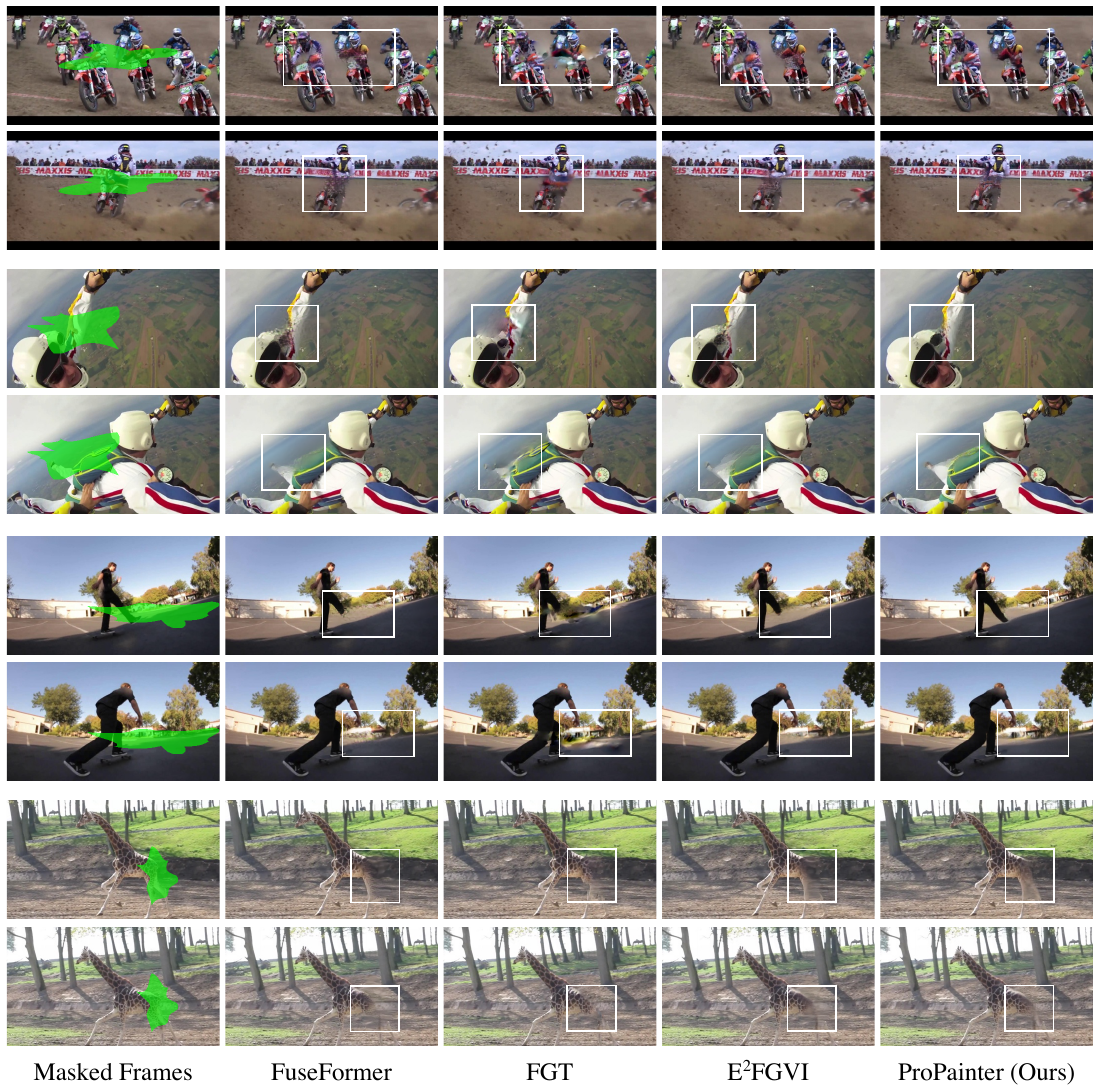}
   \caption{Qualitative comparisons on YouTube-VOS~\cite{xu2018youtube} dataset. Our ProPainter exhibits superiority in producing complete and faithful textures, resulting in enhanced spatiotemporal coherence for video inpainting. (\textbf{Zoom in for best view.})}
   \vspace{-2mm}
   \label{fig:visual_compare_youtube}
\end{figure}

\begin{figure}[t]
  \centering
   \includegraphics[width=0.99\linewidth]{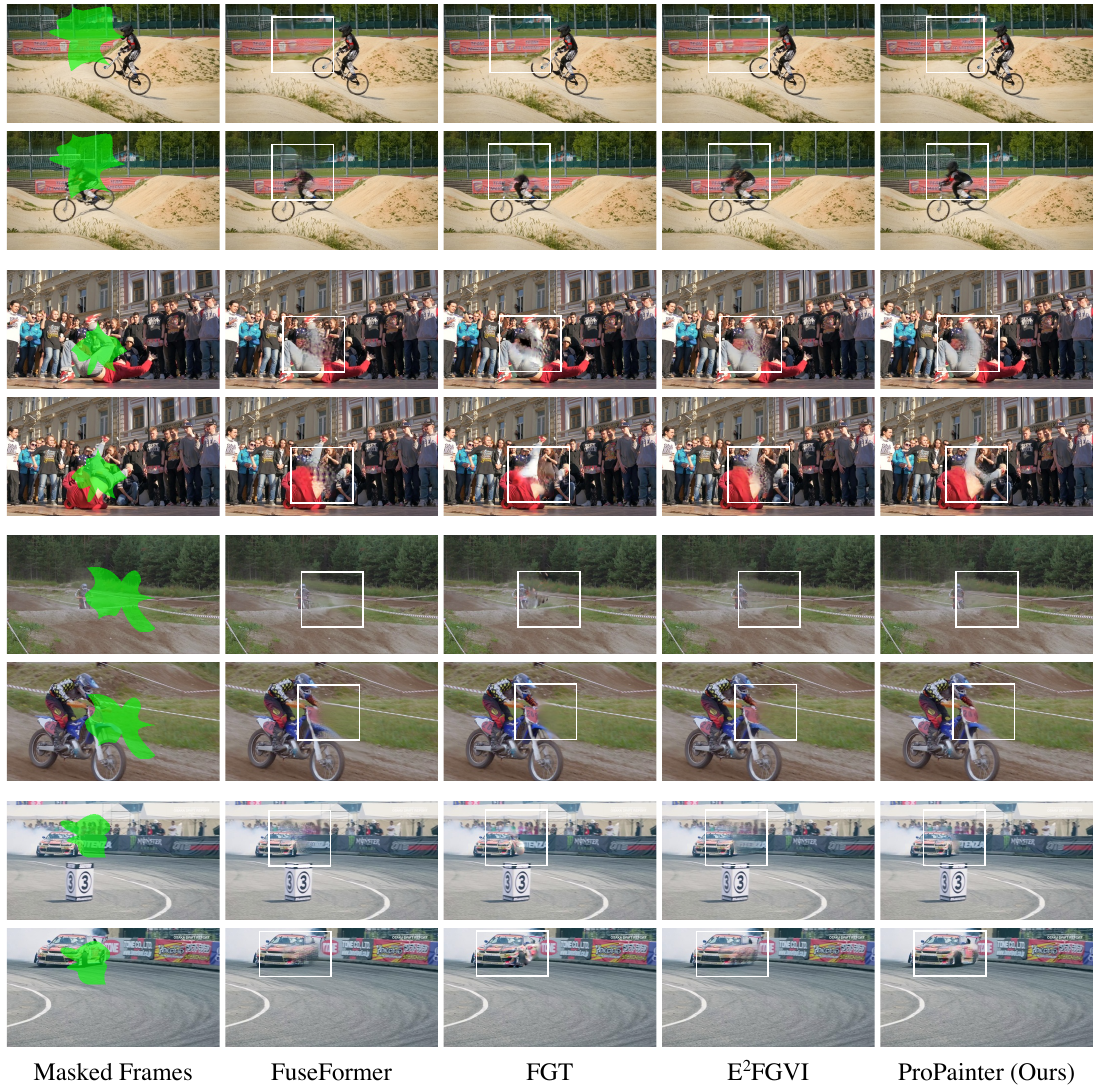}
   \caption{Qualitative comparisons on DAVIS~\cite{perazzi2016benchmark} dataset. Our ProPainter exhibits superiority in producing complete and faithful textures, resulting in enhanced spatiotemporal coherence for video inpainting. (\textbf{Zoom in for best view.})}
   \vspace{-2mm}
   \label{fig:visual_compare_davis}
\end{figure}

In this section, we provide additional visual comparisons of our method with the state-of-the-art methods, including FuseFormer~\cite{liu2021fuseformer}, FGT~\cite{zhang2022flow}, and E$^2$FGVI~\cite{li2022towards}.  
Figures~\ref{fig:visual_compare_youtube} and~\ref{fig:visual_compare_davis} present the comparisons of video completion performance on the YouTube-VOS~\cite{xu2018youtube} and DAVIS~\cite{perazzi2016benchmark} datasets, respectively.

\begin{figure}[h]
  \centering
   \includegraphics[width=0.8\linewidth]{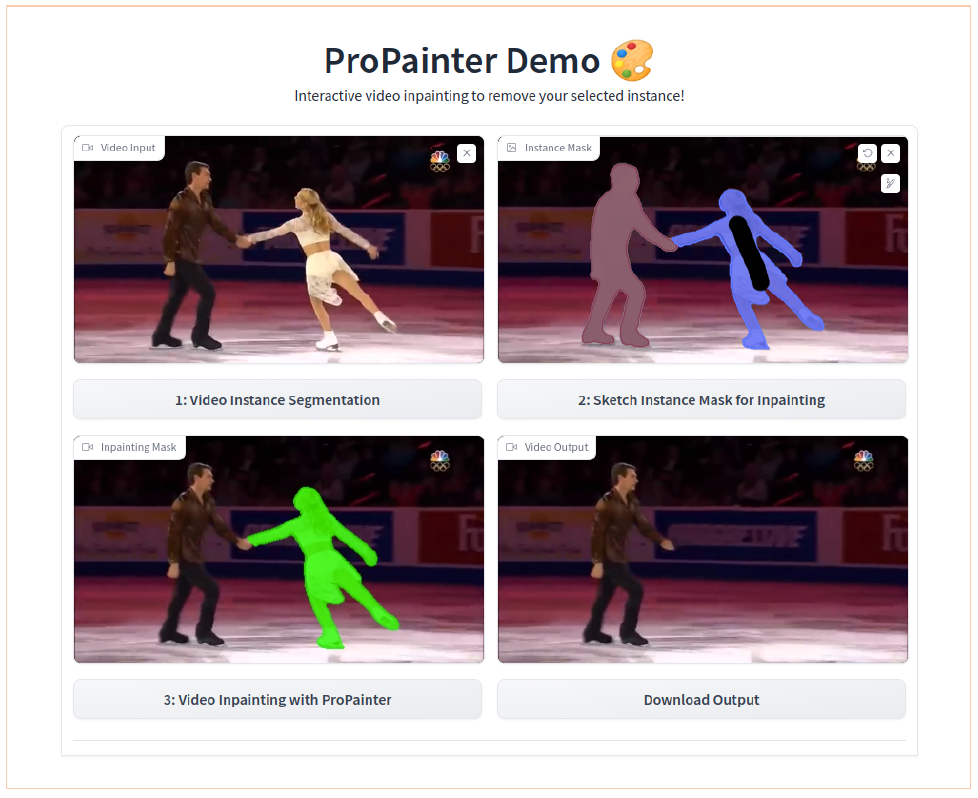}
   \caption{A screenshot of the interactive ProPainter demo.}
   \vspace{-2mm}
   \label{fig:inpainting_demo}
\end{figure}

\clearpage
Furthermore, our \href{https://shangchenzhou.com/projects/ProPainter}{[project page]} provides a video demo that showcases some results of object removal, along with an interactive demo using ProPainter. This demo incorporates a video instance segmentation network and enables users to select and remove specific objects from the video. A screenshot of this demo is presented in Figure~\ref{fig:inpainting_demo}.

\end{CJK}
\end{document}